\documentclass[11pt]{article}

\usepackage{acl}

\usepackage{times}
\usepackage{latexsym}

\usepackage[T1]{fontenc}
\usepackage[utf8]{inputenc}

\usepackage{microtype}

\usepackage{inconsolata}

\usepackage{graphicx}
\usepackage{amsmath}
\usepackage{amssymb}
\usepackage{booktabs}
\usepackage{multirow}
\usepackage{subcaption}
\usepackage{fvextra}
\usepackage{xcolor}
\usepackage[capitalize]{cleveref}
\usepackage{paralist}
\usepackage{xspace}
\usepackage{comment}

\DefineVerbatimEnvironment{CaseText}{Verbatim}{fontsize=\scriptsize,breaklines=true,breakanywhere=true,breaksymbolleft={}}

\newcommand{\heads}{\mathcal{H}}
\newcommand{\papertablefont}{\small}

\newcommand{\mmrh}{MMRetHead\xspace}
\newcommand{\mmrhs}{MMRetHeads\xspace}

\newcommand{\ie}{i.e.,\xspace}
\crefname{section}{Sec.}{Secs.}
\Crefname{section}{Sec.}{Secs.}

\title{Can Retrieval Heads See Images? Multimodal Retrieval Heads in Long-Context Vision-Language Models}

\author{
  Aaron Branson Cigres Li$^{1,*}$, Zhaowei Wang$^{1,*}$, Yu Zhao$^2$, Yiming Du$^4$ \\ \textbf{Haobo Li$^1$, Xiyu Ren$^1$, Ginny Wong$^5$, Simon See$^5$, Lishu Luo$^6$} \\ \textbf{Haodong Duan$^4$, Pasquale Minervini$^{2,3}$, Yangqiu Song$^1$} \\ 
    $^1$HKUST \qquad $^2$University of Edinburgh \qquad $^3$Miniml.AI \qquad $^4$CUHK \\
    $^5$NVAITC, NVIDIA, Santa Clara, USA \qquad
    $^6$Tsinghua University \\
    \texttt{abcli@connect.ust.hk} \qquad \texttt{\{zwanggy, yqsong\}@cse.ust.hk}
  }

\begin{document}
\maketitle

\renewcommand{\thefootnote}{\fnsymbol{footnote}}
\footnotetext[1]{Equal contribution.}
\renewcommand{\thefootnote}{\arabic{footnote}}

\begin{abstract}
Large vision-language models increasingly rely on long-context modeling to reason over documents, hour-level videos, and long-horizon agent trajectories, requiring them to locate relevant evidence across interleaved text and images.
Prior work has studied this behavior using retrieval heads in large language models, but its copy-based criterion does not directly apply when evidence appears in images.
We introduce a multimodal retrieval head detection method that scores attention from question tokens to textual or visual evidence.
With this method, we show that multimodal retrieval heads are sparse, intrinsic, and causally important: only 4.4--10.2\% of attention heads account for 50\% of the positive retrieval-score mass, and masking the top-5\% selected heads reduces performance on two tasks from MMLongBench by 42.5 and 62.3 percentage points, while random-head masking is far less damaging.
Further analysis shows that these heads are partly shared across modalities yet remain dynamic within each modality, with image retrieval heads changing more than text retrieval heads as context length and haystack modality change.
Without further training, we find that these heads can also be used directly to rank visually rich documents: on MMDocIR, Qwen3-VL-8B selected-head scoring improves Recall@1 by 7.7/7.4 macro/micro points for page retrieval and 6.3/6.8 points for layout retrieval over the strongest reported baseline.\footnote{
Code and data are available at \url{https://github.com/HKUST-KnowComp/MMRetHead}.
}
%
%Overall, these findings establish multimodal retrieval heads as practical retrieval mechanisms in LVLMs.
%
\end{abstract}

\section{Introduction}

\begin{figure}[t]
\centering
\includegraphics[width=\linewidth]{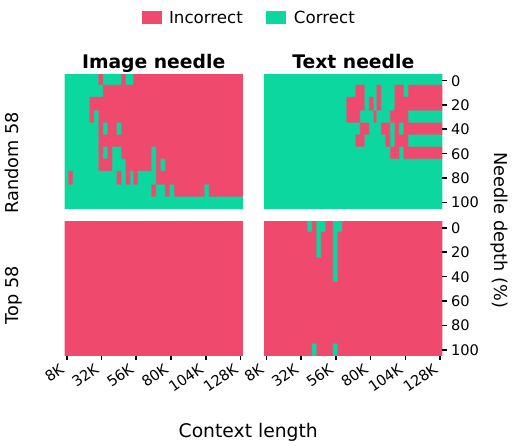}
\vspace{-0.3in}
\caption{
Removing multimodal retrieval heads causally disrupts MM-NIAH text- and image-needle retrieval.
We evaluate Qwen3-VL-8B~\citep{qwen3vl2025} on MM-NIAH~\citep{mmniah2024} across multiple context lengths and needle depths.
}
\label{fig:causal}
\vspace{-10pt}
\end{figure}

% Start of intro
Recent long-context vision-language models (LCVLMs) extend the context window to interleaved text-image inputs with tens to hundreds of thousands of tokens \citep{bytedanceseed2024seed21,seed2026seed2,gemma3technicalreport}, enabling applications over long visually rich documents~\citep{mmlongbench2025}, hour-level videos~\citep{fu2025video}, and long-horizon agent trajectories~\citep{geng2025webwatcher}.
Effective use of such long multimodal contexts requires locating task-relevant evidence across interleaved text and images.
In large language models (LLMs), this evidence-location behavior has been studied through retrieval heads, where~\citet{wu2025retrievalheads} identifies attention heads that copy answer spans from needle token-by-token during Needle-in-a-Haystack tasks~\citep[NIAH,][]{kamradt2023needle}.
However, in multimodal prompts, the evidence may be encoded visually rather than as prompt text, such as in charts, tables, or photos. The LVLMs may therefore use visual evidence to generate the answer without copying any text span from the prompt, so the copy-paste criterion may not directly apply in this setting.
To identify retrieval heads in LVLMs, we introduce multimodal retrieval heads (\mmrhs), defined as attention heads that have a high average question-to-evidence attention mass.
For each head, we compute this score by summing the post-softmax attention from each question token to all tokens within the annotated needle spans, and then averaging the resulting mass over question tokens (\cref{sec:attention-retrieval-score}).
These evidence spans may correspond to text tokens or images mapped to visual tokens.
% 
% We use the MMLongBench~\citep{mmlongbench2025} version of MM-NIAH~\citep{mmniah2024} as our detection dataset, which standardizes context lengths for controlled evaluation.
We use the MM-NIAH~\cite{mmniah2024} task from MMLongBench ~\citep{mmlongbench2025} to detect multimodal retrieval heads, which comprises four distinct tasks (\ie text retrieval, image retrieval, rendered-text retrieval, and identical-image retrieval) with standardized context lengths for controlled experiments (\cref{sec:detection-data}).
Each example consists of a long interleaved text-image haystack containing many distractor passages or images and a single question-relevant needle, the evidence item needed to answer the question. 
%
% We detect multimodal retrieval heads using four MM-NIAH tasks: \textit{1)} text retrieval: finding a text needle in a haystack; \textit{2)} image retrieval: matching candidate images to an overlaid visual needle; \textit{3)} rendered-text retrieval: locating text embedded in an image; and \textit{4)} identical-image retrieval: verifying a specific image's presence within the haystack.
%

% 
We characterize the behavior of retrieval heads in LVLMs and identify the following core properties. \textit{Sparsity} (\cref{sec:properties}): we find that a very small fraction of heads accounts for the vast majority of the retrieval score mass. \textit{Causality} (\cref{sec:causality}): ablating these sparse heads severely degrades multimodal long-context retrieval and downstream reasoning, causing the model to hallucinate evidence or explicitly state that the needed context is missing, while zeroing out random heads has minimal impact. \textit{Preservation} (\cref{fig:preservation}): we show that high-scoring multimodal retrieval heads are largely identical to those in their respective base models, suggesting these mechanisms are inherited from pretraining rather than developed during subsequent vision-language adaptation. \textit{Modality specificity and dynamic adaptation} (\cref{sec:modality}): we find that text and image modalities share a common set of retrieval heads, while also relying on distinct, modality-specific ones; this division of labor dynamically adapts to the evidence format, for example, OCR-like evidence in rendered-text retrieval predominantly recruits text-retrieval structures despite the input being visually encoded.

We further demonstrate the causality and utility of our identified multimodal retrieval heads by applying them to solve multimodal document retrieval tasks (\cref{sec:retrieval}).
%
% MMDocIR is a visually rich document retrieval benchmark in which queries must be matched to supporting evidence within long documents.
%
We evaluate both page retrieval, where candidates are document pages and the goal is to rank the page containing the evidence, and layout retrieval, where candidates are finer-grained regions within pages, and the goal is to rank the evidence region.
For each candidate page or layout region, we aggregate question-to-candidate attention through the retrieval heads by summing post-softmax attention from question tokens to candidate tokens, averaging over question tokens, and then averaging over heads.
Using this score to rank candidates, the retriever outperforms strong reported baselines on both tasks.
%

%
\begin{comment}
\paragraph{Contributions}
%
In this work,
%
\begin{inparaenum}[(i)]
%
\item we introduce a multimodal retrieval head detection method that scores question-to-evidence attention over both textual and visual needle spans, with null-question calibration (\cref{sec:attention-retrieval-score});
%
\item we characterize the resulting \mmrhs and show that they are sparse (4.4--10.2\% of heads carry 50\% of the positive retrieval-score mass), largely preserved through vision-language adaptation, and causally important, with masking collapsing MM-NIAH retrieval and substantially degrading MMLongBench-Doc, SlideVQA, MMMU, MMMU-Pro, MathVision, and MathVista, while same-count random masking is far less damaging (\cref{sec:properties,sec:causality});
%
\item we show that \mmrhs are partly modality-specific and dynamic, with image-retrieval heads shifting more than text-retrieval heads under context-length and haystack-composition shifts (\cref{sec:modality}); and
%
\item we show that, without further training, the selected heads can be reused as a relevance signal for visually rich document retrieval, improving Recall@1 on MMDocIR by 7.7/7.4 macro/micro points for page retrieval and 6.3/6.8 points for layout retrieval over the strongest reported baseline (\cref{sec:retrieval}).
%
\end{inparaenum}
\end{comment}
%

% Reminders: Don't assume reader knows words, Don't make up words, Define words properly
% Take note of query/question and needle/evidence appropriate usage

\section{Related Work}
\vspace{-5pt}
\paragraph{Retrieval heads in language models.}
Prior work shows that attention heads can implement various language-model functions, including in-context completion via induction heads \citep{olsson2022incontextlearning} and repeated-token modulation via copy-suppression heads \citep{mcdougall2023copysuppression}.
Recent work~\citep{wu2025retrievalheads} detects retrieval heads using copy-paste behavior (i.e., literal token matching) during answer generation.
However, such behavior may not generalize to broader long-context tasks beyond literal matching~\citep{modarressinolima}; other work therefore detects retrieval heads directly from attention scores~\citep{zhang2025qrhead}.
In this work, we study whether similar attention-score-based retrieval heads can be detected in LVLMs when the needle is visual.
\vspace{-4pt}
\paragraph{LVLM attention and interpretability.}
For LVLMs~\citep{seed2026seed1,xiaomi2025mimo}, attention scores have been used to analyze cross-modal alignment, grounding, and hallucination \citep{aflalo2022vlinterpret,huang2024opera,wu2025flmm}, though raw attention is not by itself a causal explanation \citep{jain2019attention,serrano2019attention}.
% kang2025fewheads ,bi2025unveiling,
In this work, we study retrieval heads in LVLMs by examining whether specific attention heads concentrate attention from task-question tokens to annotated multimodal needles.

\vspace{-4pt}
\paragraph{Multimodal long-context and retrieval.}
Recent work on long-context vision-language models~\citep{wang2026training} has introduced comprehensive evaluations spanning multimodal document VQA~\citep{mmlongbench2025}, multimodal NIAH~\citep{mmniah2024}, and long-term memory~\citep{ren2026memlens}, such as MMLongBench~\citep{mmlongbench2025}.
For multimodal document VQA, prior work often uses retrieval modules to score candidate pages or passages~\citep{robertson2009probabilistic,karpukhin2020dpr,izacard2022contriever,khattab2020colbert,ma2024dse,faysse2024colpali}.
In contrast, we ask whether an LVLM's multimodal retrieval heads can directly rank candidate pages or layout units.

\section{Detecting Multimodal Retrieval Heads}
\vspace{-5pt}
\begin{figure*}[t]
\centering
\includegraphics[width=\textwidth]{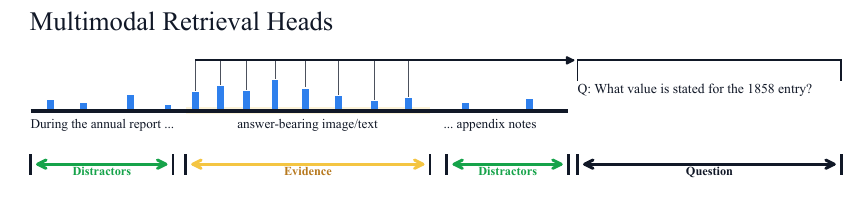}
\vspace{-0.3in}
\caption{Schematic of \mmrh detection. For each attention head, we score the post-softmax attention from question tokens to annotated task-relevant evidence spans. These evidence spans may correspond to text tokens or visual tokens, allowing the scoring framework to detect retrieval heads across multimodal inputs.}
\label{fig:method-schematic}
\end{figure*}

\subsection{Detection data} \label{sec:detection-data}
MM-NIAH \citep{mmniah2024} provides a multimodal counterpart to text-only needle-in-a-haystack tasks~\citep{kamradt2023needle}, evaluating whether LVLMs can retrieve needles from long webpage-based haystacks with interleaved text and images.
Here, we use the reprocessed version of MM-NIAH from MMLongBench \citep{mmlongbench2025}, which standardizes context lengths for tighter control across the 8K--128K settings.
In our setting, the \emph{haystack} is the full multimodal context, consisting of many distractor texts or images.
The \emph{needle} is the single task-relevant evidence inserted at a controlled depth in that context.
An LVLM is then asked a query whose answer requires retrieving this needle.
Specifically, we use text and image retrieval from MM-NIAH, covering both textual and visual needles.
In text retrieval, the needle is a short factual text snippet, such as "The beacon over the hill is a lighthouse," hidden among multimodal distractors, and the model must answer the question, such as “What is the beacon over the hill?”
%S
In image retrieval, the visual needle is an image embedded inside a larger haystack image, and the model must match it to the correct candidate option image.
We further introduce a new rendered-text retrieval task and identical image retrieval variants to separate OCR-like text retrieval and visual presence/absence verification.
%
% look out for other cases saying same but not specifying same as what. Done scanning
%
In rendered-text retrieval, the same text needle used in the text retrieval task is rendered into an image, allowing us to test whether visually encoded text recruits text-like or image-like retrieval heads. In identical image retrieval, the needle image is no longer embedded in another haystack image, and the model must determine whether it appears in the haystack, further covering OCR scenarios and augmenting the image-retrieval setting.

\subsection{Attention-Based Retrieval Score} \label{sec:attention-retrieval-score}

We define multimodal retrieval heads (\mmrhs) as attention heads with a high average question-to-evidence attention mass.

We write each example as $x=(C,q,G_x)$, where $C$ is the long interleaved text-image context serving as the haystack, $q$ is the task query, and $G_x=\{g_i\}_{i=1}^{m_x}$ is the set of task-relevant needle spans, with $g_i=(s_i,e_i)$ denoting a token interval in the tokenized context.
As shown in \cref{fig:method-schematic}, we define an attention-based retrieval score for each attention head by aggregating attention from question tokens to tokens inside all task-relevant evidence spans:
\begin{equation}
S_h(x)
=
\frac{1}{|q|}
\sum_{t_q \in q}
\sum_{g_i \in G_x}
\sum_{t=s_i}^{e_i}
A^h_{t_q \rightarrow t},
\end{equation}
\noindent where $A^h$ denotes the model's post-softmax attention score for attention head $h$. 

Then, we average the retrieval score for each attention head over all examples.
We use this formula following prior work on QRHead~\citep{zhang2025qrhead}, but adapt it to multimodal long-context inputs and show that it can identify \mmrhs in subsequent sections.

\subsection{Null-Question Calibration}

In practice, \citet{chen2025attentionrerankers} show that raw attention scores may contain question-independent biases, with certain tokens or input positions receiving high attention scores regardless of the question.
Following this work, we also use null-question calibration to measure the change in attention caused by the actual question.

Specifically, we keep the same haystack and evidence spans as the original example, but replace the task question with an uninformative null question $q_{\varnothing}$, implemented as the fixed text string ``N/A'' in the same prompt slot.
For $x=(C,q,g)$, $S^{\mathrm{null}}_h(x)$ denotes the score computed on $x_{\varnothing}=(C,q_{\varnothing},g)$.
Then, the calibrated score is computed as $S^{\mathrm{cal}}_h(x) = S_h(x) - S^{\mathrm{null}}_h(x)$.

We additionally evaluate the effect of alternative null-prompt wordings and find that they have minimal impact on the resulting head rankings; full results are reported in Appendix~\ref{app:head-score-analysis}.

\subsection{Implementation Details} \label{sec:head-selection}

We detect \mmrhs across six LVLMs: Qwen2.5-VL (7B, 32B)~\citep{qwen25vl2025}, Qwen3-VL (8B, 32B)~~\citep{qwen3vl2025}, Gemma3 (12B, 27B)~\citep{gemma3technicalreport}.
For each model, we run the four MM-NIAH tasks at five context lengths: 8K, 16K, 32K, 64K, and 128K.
Each detection condition uses 20 unique question-evidence examples, with each item evaluated at 6 needle-depth positions. Each detection condition therefore contains 120 evaluation instances. We provide full details of the sample-size stability tests in Appendix~\ref{app:detection-stability}.
Attention heads are ranked by calibrated retrieval scores, and the top 5\% of heads are selected for the following experiments. Selecting the top 5\% follows the ratio reported in prior work~\citep{wu2025retrievalheads}.
Since Gemma3 models~\citep{gemma3technicalreport} use hybrid attention with sliding-window attention, we consider only global-attention layers that can perform long-range retrieval.
\section{Basic Properties of \mmrhs} \label{sec:properties}
We first test whether \mmrhs exhibit three basic properties established in prior retrieval head work: sparsity, intrinsicity, and dynamic activation  \citep{wu2025retrievalheads}. 
%
%\subsection{Sparsity}
\paragraph{Sparsity}
We define a sparsity metric as the minimum fraction of heads required to explain 50\% of the positive calibrated retrieval score mass.
Calibrated scores can be negative when the null question attends more to the evidence spans than the task question, so this metric focuses on positive excess question-induced attention.
The score-mass sparsity metric is:
\begin{equation*}
\rho_{0.5}
=
\frac{1}{|\heads|}
\min \left\{
k :
\sum_{i=1}^{k} s_{(i)}
\ge
0.5
\sum_{h \in \heads} s_{h}
\right\},
\end{equation*}
\noindent where $\heads$ is the full model attention-head set and $s_h=\max(\bar{s}_h,0)$. $\bar{s}_h$ is the calibrated detection-set average, and $s_{(1)}\ge \cdots \ge s_{(|\heads|)}$ are sorted positive score masses.

\begin{figure}[t]
\centering
\includegraphics[width=0.9\linewidth]{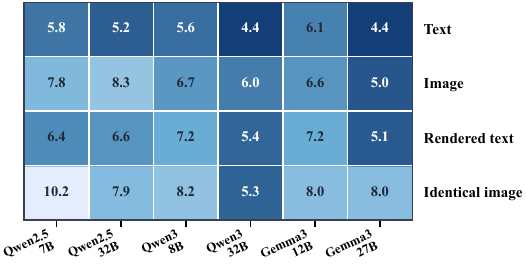}
\vspace{-0.1in}
\caption{
Retrieval-score mass concentration. Each cell shows the percentage of heads needed to cover 50\% of positive calibrated retrieval mass; darker cells indicate stronger concentration.
}
\vspace{-10pt}
\label{fig:sparsity-mass}
\end{figure}

\begin{figure}[t]
\centering
\includegraphics[width=0.85\columnwidth]{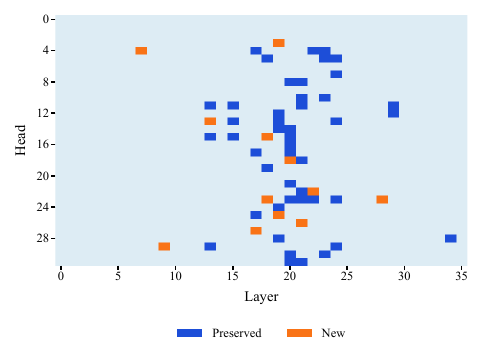}
\vspace{-0.2in}
\caption{
Top-5\% text-retrieval heads in Qwen3-VL-8B. Blue heads are preserved from Qwen3-8B, while orange heads newly enter the top-5\% set after vision-language training. We also show the effect of SFT on Gemma3-12B in Appendix~\ref{app:gemma-intrinsic}.
}
\label{fig:preservation}
\vspace{-10pt}
\end{figure}

%\cref{fig:sparsity-mass} shows that calibrated retrieval-score mass is concentrated in a small subset of attention heads across tested LVLMs and MM-NIAH tasks. Averaged over all context lengths, 4.4--10.2 percent of all model attention heads account for 50\% of the positive score mass. \zhaowei{add a conclusion, something like: so, we find that the multimodal retrieval heads remain sparse.}
%
\cref{fig:sparsity-mass} confirms that \mmrhs are sparse: averaged over all context lengths, only 4.4--10.2\% of attention heads account for 50\% of the positive score mass across tested LVLMs and MM-NIAH tasks.
Additional Head Score distributions and descriptive statistics are reported in Appendix~\ref{app:head-score-analysis}.
%
%
%
%\subsection{Intrinsicity}
\paragraph{Intrinsicity}
\citet{wu2025retrievalheads}  show that retrieval heads are largely established during pretraining, with later SFT producing only minor changes.
Here, we test both Qwen3-VL-8B and Gemma3-12B to verify whether this claim still holds for LVLMs.
For Qwen3, we test whether text retrieval heads are preserved during multimodal adaptation by comparing Qwen3-8B \citep{qwen3technicalreport} with Qwen3-VL-8B \citep{qwen3vl2025}.
For Gemma3, we test intrinsicity more directly by comparing pretrained Gemma3-12B-PT with instruction fine-tuned Gemma3-12B-IT \citep{gemma3technicalreport}.
To support text-only Qwen3-8B, we use 128 examples from Natural Questions~\citep{kwiatkowski2019natural} following the QRHead~\citep{zhang2025qrhead}.
\cref{fig:preservation} and \cref{fig:gemma-preservation} show that high-scoring text retrieval heads remain concentrated at the same layer/head locations: 46 of 58 Qwen3-VL-8B top-5\% heads and 34 of 39 Gemma3-12B-IT top-5\% heads are shared with their preceding versions. This high intersection indicates that text retrieval heads are intrinsic, being preserved through vision-language adaptation in Qwen3-VL-8B and instruction tuning in Gemma3-12B rather than newly formed during later adaptation.

\begin{figure}[t]
\centering
\includegraphics[width=0.85\linewidth]{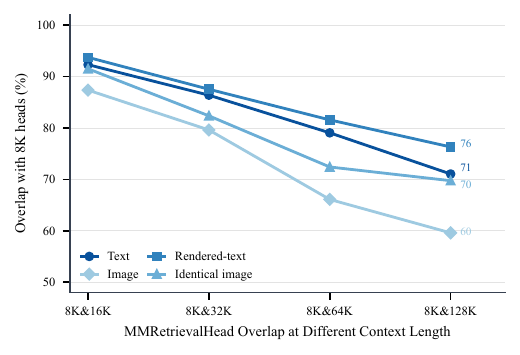}
\vspace{-0.2in}
\caption{
Context-length sensitivity of retrieval-head selection. Curves show the overlap between top-5\% heads selected at 8K and those selected at longer contexts, averaged over six LVLMs. We find that image-retrieval heads are less stable.
}
\vspace{-22pt}
\label{fig:context-length-overlap}
\end{figure}

\begin{figure}[t]
\centering
\includegraphics[width=0.9\linewidth]{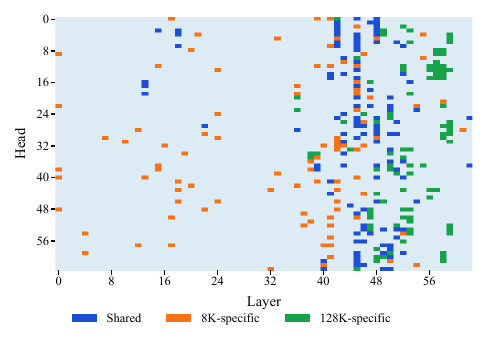}
\vspace{-5pt}
\caption{
Layer distribution of Qwen3-VL-32B image-retrieval heads at 8K and 128K. Blue heads are shared; orange and green mark 8K- and 128K-specific heads; context-specific heads show different layer distributions.
}
\vspace{-10pt}
\label{fig:retrieval-head-distribution}
\end{figure}

% preservation_qwen3_overlap_map_gemma.pdf is in appendix
\paragraph{Dynamic activations}
Prior work~\citep{wu2025retrievalheads} has shown that retrieval heads are dynamically activated, with some heads appearing only in specific contexts and others appearing consistently across contexts. Here, we examine context length as one factor driving this variation in \mmrhs. \cref{fig:context-length-overlap} shows a substantial decrease in intersection heads detected in short and long-context settings. At 128K, only 60\% of image retrieval heads, 70\% of identical-image retrieval heads, 71\% of text retrieval heads, and 76\% of rendered-text retrieval heads remain shared with the head set detected at the 8K setting. This suggests that retrieval head selection is context-length sensitive, with image retrieval heads showing the largest change across lengths.
Furthermore, \cref{fig:retrieval-head-distribution} shows a clear layer-distribution change between heads detected only at 8K setting and heads detected only at 128K setting of image retrieval. The heads unique to the 8K setting are concentrated more in earlier layers, with a mean layer of 33.1, whereas the heads unique to the 128K setting shift toward later layers, with a mean layer of 52.4. This suggests that shorter-context retrieval can rely more on earlier representations, while longer-context retrieval may require representations that have undergone more layer-wise processing.

% Conclusion and important numbers included 
%\zhaowei{DONE: move this observation to section 4 and use a title like "dynamic activation," or "image retrieval heads are more dynamic." Also, image retrieval head has a lower overlap than the text retrieval head. This is very important.}

\section{Causal Role in Multimodal NIAH, Long-Context, and Reasoning Tasks}
\label{sec:causality}

We remove the detected attention heads from LVLMs by applying a zero mask to their post-softmax attention weights, and evaluate whether retrieval performance degrades.  We apply the mask during both prompt prefill and answer decoding because this setting causes much stronger degradation than decode-only masking, indicating that retrieval also occurs while the model processes the haystack and question before generation. We provide more detail of this in Appendix~\ref{app:prefill-decode-masking}. We mainly present results on Qwen3-VL-8B. We first intervene in MM-NIAH, where synthetic examples provide known evidence spans and controlled context lengths, then test whether the detected heads remain important beyond the detection setting by masking them on Long Document VQA and multimodal reasoning benchmarks.

\subsection{Head Masking on MM-NIAH}
\label{sec:causal-controlled}

\paragraph{Causal effects on MM-NIAH.} 

\cref{fig:causal} summarizes MM-NIAH causal masking results for Qwen3-VL-8B across context lengths and evidence depths. For each task, we mask the top 5\% retrieval heads detected at the 128K context length, which corresponds to 58 heads for Qwen3-VL-8B.

We use the top-5\% masking threshold because masking only the top 1\% retains considerable retrieval capability, whereas masking the top 10\% produces substantially more repetitive outputs, in which the model repeats the same phrase or characters until the token limit. Complete results and examples are reported in Appendix~\ref{app:additional-masking-controls}.

As a control, we mask 58 randomly chosen attention heads. In both text and image retrieval settings, masking retrieval heads causes severe retrieval failures across all context-length and needle-depth settings, showing that the detected heads are causally important rather than only correlated with evidence locations.

To further verify that the causal effect is specific to the detected heads, we additionally use a layer-matched control in which random heads are sampled to match the layer distribution of the detected retrieval heads. This control remains less degrading than detected-head masking. Following prior information-flow analysis \citep{wang-etal-2023-label}, we also perform a path-specific intervention that selectively blocks attention paths from evidence tokens to question and decoded-response tokens while leaving the remaining computation of the detected heads intact. This path-specific masking also causes substantial degradation on both text and image retrieval, indicating that the detected heads mediate retrieval-related information flow rather than only general computation. Full results are reported in Appendix~\ref{app:additional-masking-controls}.

% Remove next rank from figure, keep decoding only and with prefill different. Add conclusion that it is causal

\paragraph{Cross-length causal transfer.}

\begin{figure}[t]
\centering
\includegraphics[width=\linewidth]{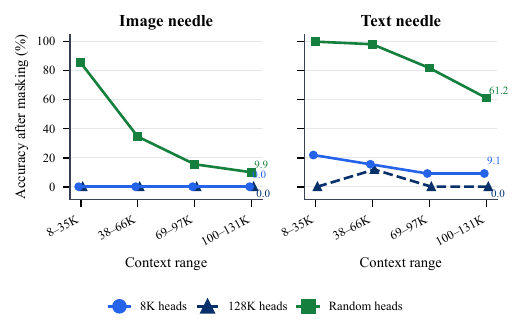}
\vspace{-0.3in}
\caption{
Cross-length causal transfer. Masking heads detected at 8K or 128K degrades MM-NIAH performance across other context lengths, showing that retrieval heads remain causally important beyond their detection length.
}
\vspace{-10pt}
\label{fig:short-context-causal}
\end{figure}

\cref{fig:context-length-overlap} shows that the retrieval head sets detected at different context lengths only partially intersect. We therefore ask whether this context-length variation weakens their causal role, or whether heads detected at one length remain functionally important at other lengths. \cref{fig:short-context-causal} shows that the causal effect transfers in both directions.
On the long context tasks, from 100K to 131K, masking either the 8K-detected heads reduces image retrieval accuracy to 0.0\%, and reduces text retrieval accuracy to 9.1\%.
Conversely, on the shorter context tasks (8K to 35K), masking the 128K-detected heads degrades both image and text retrieval accuracy to 0.0\%.
These results show that, although the selected head sets vary with context length, heads detected at one length can remain causally important at other lengths.

\subsection{Long-Document VQA Masking}
\label{sec:causal-long-context}

\begin{figure}[t]
\centering
\includegraphics[width=0.90\linewidth]{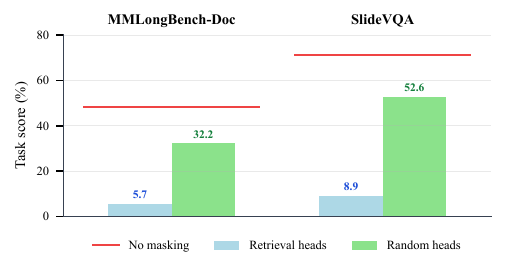}
\vspace{-0.2in}
\caption{
Causal effect on 128K Long Document VQA category of MMLongBench. Masking retrieval heads detected on 128K MM-NIAH degrades MMLongBench-Doc and SlideVQA far more than random masking.
}
\label{fig:mmlongbench-masking}
\vspace{-10pt}
\end{figure}

We next evaluate Long Document VQA using the 128K context-length MMLongBench-Doc and SlideVQA subsets from MMLongBench~\citep{mmlongbench2025}. We mask the union of retrieval heads detected from the four MM-NIAH tasks introduced in \cref{sec:detection-data}, using their 128K detection settings. \cref{fig:mmlongbench-masking} shows that masking these retrieval heads sharply reduces MMLongBench-Doc score from 48.2\% to 5.7\% and SlideVQA score from 71.2\% to 8.9\%. In contrast, masking randomly chosen attention heads is less damaging, leaving scores of 32.2\% and 52.6\%, respectively. These results show that retrieval heads identified in controlled MM-NIAH tasks remain causally important on Long Document VQA tasks beyond the detection setting.

\subsection{Reasoning Benchmark Masking}
\label{sec:downstream}

For broader downstream masking, we evaluate Qwen3-VL-8B on MMMU~\citep{yue2024mmmu}, MMMU-Pro~\citep{yue2024mmmupro}, MathVision~\citep{wang2024mathvision}, and MathVista~\citep{lu2024mathvista}. We mask the union of retrieval heads detected from the four 128K MM-NIAH task settings in \cref{sec:detection-data}. We also compare direct-answer prompting with chain-of-thought (CoT) prompting~\citep{wei2022chainofthought}.

\begin{figure}[t]
\centering
\includegraphics[width=\linewidth]{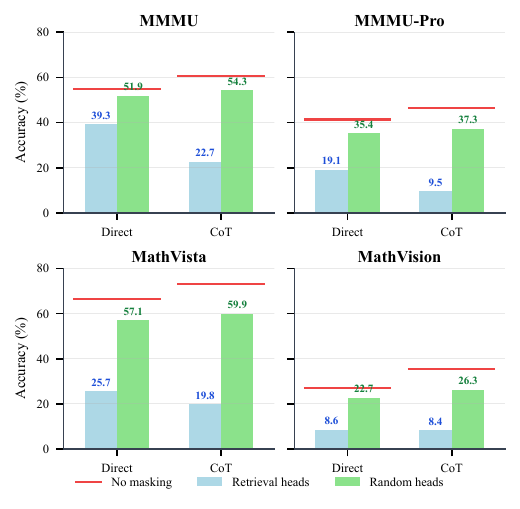}
\vspace{-0.3in}
\caption{
Qwen3-VL-8B accuracy on multimodal reasoning. We compare no masking, retrieval head masking, and random head masking across direct-answer and CoT prompting. Masking our retrieval heads causes substantially larger accuracy drops.
}
\vspace{-20pt}
\label{fig:downstream-masking}
\end{figure}

As shown in \cref{fig:downstream-masking}, masking \mmrhs causes much larger accuracy drops than masking randomly chosen heads. Across the four benchmarks, direct-answer accuracy drops by an average of 24.3 points, while CoT accuracy drops by an average of 38.8 points, showing that heads detected on controlled MM-NIAH tasks remain causally important for downstream multimodal reasoning. Furthermore, the larger CoT drop suggests that multi-step reasoning relies more heavily on repeated grounding, though it may also reflect longer masked-decoding trajectories.
With CoT prompting, the model outputs intermediate reasoning, allowing us to inspect failure modes when retrieval heads are masked. We find several recurring failures. Some failures indicate disrupted access to the relevant evidence. For example, in an MMMU-Pro thermodynamics problem, the masked model states that there is ``Not enough information provided'' even though the needed diagram is included in the prompt. In other cases, it hallucinates supporting content or extracts the wrong information. Prompt-output examples are provided in Appendix~\ref{app:downstream-failure-cases}.

\section{Modality-Specific Retrieval Head} %
\label{sec:modality}

We test whether \mmrhs are shared across retrieval tasks by measuring the intersection between the detected head sets for text and image retrieval, and for text and rendered-text retrieval.
For two top-5\% head sets $\heads_A$ and $\heads_B$ of equal size, we report their intersection as $|\heads_A \cap \heads_B| / |\heads_A|$, i.e., the fraction of heads in one set that also appear in the other.
The ratio is symmetric in $A$ and $B$, since $|\heads_A| = |\heads_B|$, 

We then test whether retrieval heads are sensitive to the modality ratio of the haystack. For this analysis, we use the MM-NIAH text retrieval and image retrieval tasks, preserve the answer-bearing needle entries, and vary only haystack. Text are grouped into units of esimated 500 tokens using a character-length heuristic, where one token is approximated four characters, while each image is treated as one unit. Vision-token counts are model-dependent: Gemma3 uses roughly 256 vision tokens per image, while Qwen3-VL and Qwen2.5-VL use dynamic image tokenization, with averages of about 209 and 270 vision tokens per image in our sampled composition runs, respectively. We then sample 128 units per example and set the target image-unit ratio to 0, 0.1, 0.2, or 0.4, corresponding to 0, 13, 26, or 51 image units out of 128.

% remove the line in figure for image/image existing retrieval

\paragraph{Text and image retrieval use partially distinct heads.}

\begin{figure}[t]
\centering
\includegraphics[width=\linewidth]{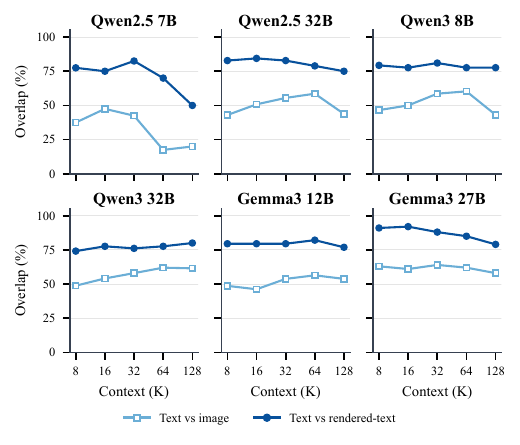}
\vspace{-0.3in}
\caption{
Retrieval-task sensitivity of retrieval-head selection at context lengths from 8K to 128K. Light-blue curves compare text and image retrieval tasks, while dark-blue curves compare text and rendered-text retrieval; we find that rendered-text retrieval consistently overlaps more with text retrieval.
}
\vspace{-10pt}
\label{fig:text-modality-overlap}
\end{figure}

\cref{fig:text-modality-overlap} shows that text and image retrieval heads are partially shared but not identical. Across six LVLMs and five context lengths, the intersection between the top-5\% text and image retrieval head sets ranges from 0.18 to 0.64, with an average intersection of 0.51. This suggests both modality-agnostic and modality-specific retrieval heads.

% is this direct answer already?

\paragraph{Rendered-text retrieval preserves text-retrieval structure.}

\cref{fig:text-modality-overlap} also shows that rendered-text retrieval is much closer to text retrieval than to ordinary image retrieval. Across six LVLMs and five context lengths, text/rendered-text head intersection ranges from 0.5 to 0.92, with an average intersection of 0.79. This is much higher than the 0.51 average intersection between text and image retrieval. This suggests that retrieval heads are sensitive not only to the raw input channel, but even more to the content being retrieved.

\paragraph{Haystack text-image ratio induces modality-aware head dynamics.}

\begin{figure}[t]
\centering
\includegraphics[width=\linewidth]{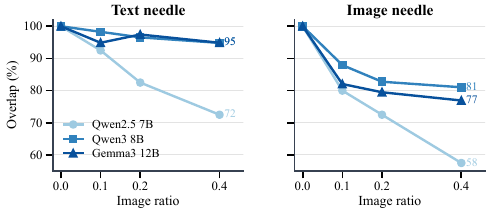}
\vspace{-0.3in}
\caption{Sensitivity to the haystack image ratio. Each curve shows the intersection between the top-5\% heads detected at a given image ratio and those detected in the all-text haystack setting. Both text and image retrieval heads change as the image ratio increases.}
\label{fig:haystack-composition}
\vspace{-10pt}
\end{figure}

\cref{fig:haystack-composition} shows that retrieval head selection is sensitive to the text-image ratio of the haystack.
As images make up a larger share of the haystack, with the image ratio increasing from 0 to 0.4, the intersection with the head set detected in the all-text haystack decreases for both text and image retrieval.
The decrease is much larger for image retrieval.
When the image ratio is 0.4, the mean intersection across the three models is 0.87 for text retrieval, and 0.72 for image retrieval. 
This pattern suggests that the detected heads depend not only on the target evidence modality but also on the haystack's modality.
For image retrieval, the stronger decrease in intersection suggests that the model recruits more modality-specific heads when haystack images create additional visual noise.

\section{Multimodal Document Re-Ranking}
\label{sec:retrieval}

We apply the selected retrieval heads to multimodal document re-ranking, where the goal is to rank candidate pages or layout regions by whether they contain the evidence needed to answer a question, and find that this method outperforms strong baselines.

\paragraph{Method and Calibration}

Given a selected retrieval head set $\heads_{\mathrm{sel}}$, a question $q$, and candidate units $D=\{d_i\}_{i=1}^{n}$, we compute a retrieval score by aggregating question-to-candidate attention through the selected heads.

For re-ranking, $\heads_{\mathrm{sel}}$ is the union of the top-5\% head sets detected on the four MM-NIAH tasks: text retrieval, image retrieval, rendered-text retrieval, and identical-image retrieval.

We define the retrieval head score $R(d_i \mid q; \heads_{\mathrm{sel}})$ for candidate units $d_i$ as:
\begin{equation}
R = \frac{1}{|\heads_{\mathrm{sel}}|}
\sum_{h \in \heads_{\mathrm{sel}}}
\frac{1}{|q|}
\sum_{t_q \in q}
\sum_{t_d \in d_i}
A^{h}_{t_q \rightarrow t_d}.
\end{equation}
We also apply the null-question calibration in this experiment.
Candidate units are then ranked using the null-calibrated retrieval score as a relevance metric \citep{chen2025attentionrerankers}. 

\paragraph{Re-Ranking Performance}

Multimodal document re-ranking requires retrieving evidence from visually rich documents whose evidence may appear as rendered text, images, or layout-dependent structure.
For example, in a manual, answering a question may require interpreting an illustrated step, while in a research report, the relevant evidence may be a reported percentage in a chart. 

To construct a retriever, we use the selected retrieval heads in Qwen3-VL-8B and Gemma3-12B, detected at 128K context length. We evaluate our LVLM-based retriever on MMDocIR \citep{dong2025mmdocir} at both the page and layout levels. Page-level retrieval ranks document pages, while layout-level retrieval ranks detected layout elements within pages, such as text blocks, headings, equations, tables, figures, or charts, using bounding-box evidence annotations. To keep inputs within the LVLM context window, we cap page retrieval at 200 pages per forward pass, scoring each 200-page group separately for longer documents; for layout retrieval, we cap each forward pass at 50 images.

As shown in \cref{tab:mmdocir-summary}, the retriever built from Qwen3-VL-8B selected retrieval heads achieves the best MMDocIR performance among the compared methods at both page and layout levels.\footnote{Recall@1 measures whether the top-ranked candidate contains the annotated evidence; macro Recall@1 averages across domains, micro across all questions.} For page retrieval, the retriever reaches 64.7 macro and 64.5 micro Recall@1, improving over the strongest reported baseline, Col-Phi3, by 7.7 and 7.4 percentage points, respectively. For layout retrieval, it reaches 39.0 macro and 39.3 micro Recall@1, improving over the strongest reported baseline, ColPali, by 6.3 and 6.8 percentage points. Full Recall@k results are provided in \cref{tab:mmdocir-full}. These results show that attention from the selected retrieval heads provides a usable relevance signal for realistic multimodal document retrieval.

Finally, to test whether the gains come from selected retrieval heads rather than generic LVLM attention, we add a baseline where all attention heads are used to compute retrieval scores. For Qwen3-VL-8B, aggregating all attention heads still outperforms the strongest listed page baseline, but selected retrieval heads improve over all-head attention by 2.9 macro Recall@1 points and 3.2 micro Recall@1 points, with smaller gains at Recall@3 and Recall@5. 

\begin{table}[t]
\centering
\papertablefont
\setlength{\tabcolsep}{6pt}
\renewcommand{\arraystretch}{0.96}
\begin{tabular}{lrrc}
\toprule
Method & \shortstack{Macro\\R@1} & \shortstack{Micro\\R@1} & \shortstack{$\Delta$ vs.\\baseline} \\
\midrule
\multicolumn{4}{c}{\textit{Page-Level}} \\
\midrule
DSE$_{\text{wiki-ss}}$ & 48.0 & 47.5 & -- \\
 DSE$_{\text{docmatix}}$ & 50.2 & 50.1 & -- \\
 ColPali & 53.0 & 52.7 & -- \\
 DPR-Phi3$_{\text{MMDocIR}}$ & 54.1 & 53.7 & -- \\
 Col-Phi3$_{\text{MMDocIR}}$ & 57.0 & 57.1 & -- \\
All-head & 61.8 & 61.2 & +4.8/+4.1 \\
 \mmrhs (ours) & \textbf{64.7} & \textbf{64.5} & +7.7/+7.4 \\
\midrule
\midrule
\multicolumn{4}{c}{\textit{Layout-Level}} \\
\midrule
DSE$_{\text{wiki-ss}}$ & 28.2 & 29.2 & -- \\
DSE$_{\text{docmatix}}$ & 27.9 & 29.1 & -- \\
ColPali & 32.7 & 32.5 & -- \\
DPR-Phi3$_{\text{MMDocIR}}$& 29.5 & 30.2 & -- \\
Col-Phi3$_{\text{MMDocIR}}$& 31.1 & 31.6 & -- \\
All-head & 36.3 & 37.2 & +3.6/+4.7 \\
\mmrhs (ours) & \textbf{39.0} & \textbf{39.3} & +6.3/+6.8 \\
\bottomrule
\end{tabular}

\caption{
MMDocIR Recall@1 summary. We evaluate our retriever built on Qwen3-VL-8B and report macro/micro Recall@1 for page- and layout-level retrieval, with deltas against the strongest reported baseline. We compare with thirteen strong baselines in Appendix~\ref{app:mmdocir-details}. \mmrhs achieves the best performance in both settings.
}
\vspace{-10pt}
\label{tab:mmdocir-summary}
\end{table}

\section{Conclusions}

This work shows that LVLMs contain multimodal retrieval heads, a compact set of attention heads whose question-to-evidence attention identifies relevant multimodal evidence. These heads are sparse and dependent on context length and modality, and partly shared across text and image retrieval. Masking retrieval heads causally disrupts Long Document VQA and downstream multimodal reasoning. Beyond intervention, we show that those multimodal retrieval heads are also useful for downstream tasks: aggregating selected-head attention provides a strong retriever for ranking visually rich documents. Overall, we hope this work provides a useful lens for studying multimodal long-context and inspires more interpretability study of LVLMs.

\section*{Limitations}

Our analysis uses attention scores as a proxy for retrieval-oriented heads. Masking shows strong behavioral effects, but a full circuit-level account would also need to test MLPs, residual-stream pathways, and non-selected heads.

Detected head sets may depend on context length, detection dataset, and language. The MM-NIAH tasks cover only English examples up to 128K context length. They also do not cover all real multimodal retrieval settings, such as dense documents, OCR-heavy pages, charts, diagrams, or multi-image reasoning. The analysis covers two decoder-only LVLM families with vision-token inputs, so the findings may not generalize to cross-attention LVLMs, other tokenizers, or different vision encoders.

The retrieval head-based retriever requires access to internal attention weights and can be more expensive than embedding-based retrieval because it runs LVLM forward passes over candidate context. Our evaluation does not fully test latency, memory cost, or index-time/query-time trade-offs. In settings with many repeated queries over a fixed corpus, embedding or late-interaction retrievers are likely to remain preferable. Future work should compare deployment cost and retrieval quality directly.

\section*{Ethical Considerations}
This paper utilizes several publicly available datasets, including MM-NIAH \citep{mmniah2024}, MMLongBench \citep{mmlongbench2025}, Natural Questions \citep{kwiatkowski2019natural}, MMMU \citep{yue2024mmmu}, MMMU-Pro \citep{yue2024mmmupro}, MathVision \citep{wang2024mathvision}, MathVista \citep{lu2024mathvista}, and MMDocIR \citep{dong2025mmdocir}, which are accessible to the research community under MIT, MIT, CC BY-SA 3.0, Apache 2.0, Apache 2.0, MIT, CC BY-SA 4.0, and Apache 2.0 licenses, respectively. All data are derived from previously released open-source benchmarks, and we use them solely for non-commercial research evaluation in accordance with their respective terms. The data do not contain personally identifying information beyond what is already public in the source corpora, so our work does not raise additional privacy concerns regarding specific entities.

Our experiments involve the use of Qwen2.5-VL \citep{qwen25vl2025}, Qwen3-VL \citep{qwen3vl2025}, and Gemma3 \citep{gemma3technicalreport}, released under the Apache 2.0, Apache 2.0, and Gemma Terms of Use, respectively, so the same risks from LVLM research are also applicable to this work.

\section*{Acknowledgements}
The authors of this paper were supported by the National Key Research and Development Program of China (2025YFE0200500), the ITSP Platform Research Project (ITS/189/23FP) from ITC of Hong Kong, SAR, China, and the AoE (AoE/E-601/24-N), the CRF (No. C6004-25G), the RIF (R6021-20) and the GRF (16205322) from RGC of Hong Kong, SAR, China. We also thank the NVIDIA AI Technology Center (NVAITC) for the support and additional funding.

\bibliography{references}

\clearpage
\nolinenumbers
\appendix

\section{MMDocIR Detailed Retrieval Results}
\label{app:mmdocir-details}

\subsection{Full Recall@k results}

We provide the full MMDocIR page- and layout-level Recall@k results in \cref{tab:mmdocir-full}. 

\begin{table*}[t]
\centering
\papertablefont
\setlength{\tabcolsep}{4pt}
\renewcommand{\arraystretch}{1.02}
\begin{tabular}{llrrrrrr}
\toprule
Page method & Scoring & \multicolumn{3}{c}{Macro} & \multicolumn{3}{c}{Micro} \\
\cmidrule(lr){3-5}\cmidrule(lr){6-8}
 & & R@1 & R@3 & R@5 & R@1 & R@3 & R@5 \\
\midrule
DPR & VLM-text & 27.2 & 46.3 & 57.8 & 26.9 & 46.2 & 57.8 \\
ColBERT & VLM-text & 45.8 & 64.9 & 72.3 & 44.9 & 64.8 & 72.3 \\
BGE & VLM-text & 40.6 & 59.7 & 68.4 & 39.6 & 59.6 & 68.5 \\
E5 & VLM-text & 40.8 & 60.3 & 69.1 & 39.5 & 59.3 & 67.9 \\
Contriever & VLM-text & 40.9 & 60.6 & 69.2 & 39.7 & 59.7 & 68.3 \\
GTE & VLM-text & 38.9 & 58.7 & 67.6 & 37.9 & 58.3 & 67.2 \\
DSE$_{\text{wiki-ss}}$ & Image & 48.0 & 70.6 & 78.5 & 47.5 & 71.4 & 79.2 \\
DSE$_{\text{docmatix}}$ & Image & 50.2 & 71.4 & 79.5 & 50.1 & 71.8 & 80.1 \\
ColPali & Image & 53.0 & 74.7 & 80.8 & 52.7 & 75.0 & 81.0 \\
DPR-Phi3$_{\text{MMDocIR}}$ & Image & 54.1 & 73.5 & 81.1 & 53.7 & 74.3 & 81.8 \\
Col-Phi3$_{\text{MMDocIR}}$ & Image & 57.0 & 76.3 & 82.2 & 57.1 & 76.8 & 83.0 \\
Qwen3-VL-8B all-head attention & LVLM attn. & 61.8 & 81.2 & 87.5 & 61.2 & 81.2 & 87.7 \\
Gemma3-12B all-global attention & LVLM attn. & 49.4 & 71.0 & 78.6 & 50.0 & 70.8 & 78.9 \\
\midrule
Qwen3-VL-8B selected heads & LVLM attn. & \textbf{64.7} & \textbf{83.8} & \textbf{88.9} & \textbf{64.5} & \textbf{83.6} & \textbf{88.9} \\
Gemma3-12B selected heads & LVLM attn. & 51.8 & 73.4 & 80.2 & 52.4 & 73.4 & 80.7 \\
\midrule
Layout method & Scoring & \multicolumn{3}{c}{Macro} & \multicolumn{3}{c}{Micro} \\
\cmidrule(lr){3-5}\cmidrule(lr){6-8}
 & & R@1 & R@5 & R@10 & R@1 & R@5 & R@10 \\
\midrule
DPR & VLM-text & 19.3 & 39.8 & 49.5 & 19.2 & 40.4 & 50.5 \\
ColBERT & VLM-text & 31.3 & 54.4 & 61.9 & 31.4 & 56.0 & 63.7 \\
BGE & VLM-text & 28.3 & 51.8 & 60.3 & 29.0 & 53.2 & 62.4 \\
E5 & VLM-text & 26.7 & 51.1 & 60.4 & 26.4 & 51.8 & 61.2 \\
Contriever & VLM-text & 28.3 & 51.7 & 60.4 & 28.9 & 53.0 & 62.2 \\
GTE & VLM-text & 26.1 & 51.3 & 60.7 & 27.1 & 52.3 & 62.2 \\
DSE$_{\text{wiki-ss}}$ & Image & 28.2 & 50.2 & 58.5 & 29.2 & 52.1 & 61.1 \\
DSE$_{\text{docmatix}}$ & Image & 27.9 & 49.8 & 57.5 & 29.1 & 51.9 & 59.9 \\
ColPali & Image & 32.7 & 54.0 & 62.0 & 32.5 & 54.3 & 63.2 \\
DPR-Phi3$_{\text{MMDocIR}}$ & Image & 29.5 & 51.6 & 60.2 & 30.2 & 53.9 & 62.8 \\
Col-Phi3$_{\text{MMDocIR}}$ & Image & 31.1 & 52.3 & 61.1 & 31.6 & 54.5 & 63.3 \\
Qwen3-VL-8B all-head attention & LVLM attn. & 36.3 & 64.2 & 73.3 & 37.2 & 66.3 & 75.9 \\
Gemma3-12B all-global attention & LVLM attn. & 25.6 & 51.6 & 61.8 & 24.2 & 50.9 & 63.1 \\
\midrule
Qwen3-VL-8B selected heads & LVLM attn. & \textbf{39.0} & \textbf{64.9} & \textbf{74.2} & \textbf{39.3} & \textbf{67.1} & \textbf{76.6} \\
Gemma3-12B selected heads & LVLM attn. & 28.2 & 55.2 & 64.7 & 27.6 & 55.2 & 66.1 \\
\bottomrule
\end{tabular}
\captionsetup{hypcap=false}
\captionof{table}{Full MMDocIR page and layout Recall@k results under macro domain averaging and micro query averaging. Baseline rows are reported numbers from the MMDocIR benchmark paper; selected-head and all-head rows are our runs. Bold marks the best direct retrieval or attention row.}
\label{tab:mmdocir-full}
\end{table*}

\subsection{Domain Breakdown}
\label{app:mmdocir-domain}

We provide the MMDocIR domain-level Recall@1 results in \cref{tab:mmdocir-domain-r1}.

\begin{table*}[t!]
\centering

\papertablefont
\setlength{\tabcolsep}{3pt}
\begin{tabular}{lrrrr}
\toprule
Domain &
\shortstack{Page\\best baseline} &
\shortstack{Page\\Qwen3-VL-8B} &
\shortstack{Layout\\best baseline} &
\shortstack{Layout\\Qwen3-VL-8B} \\
\midrule
Research report   & 58.9 & \textbf{61.5} & 23.4 & \textbf{36.4} \\
Admin/industry    & 51.8 & \textbf{60.3} & 22.1 & \textbf{32.2} \\
Tutorial/workshop & 58.6 & \textbf{65.1} & 37.5 & \textbf{40.9} \\
Academic paper    & 61.3 & \textbf{70.3} & 35.2 & \textbf{42.0} \\
Brochure          & 57.3 & \textbf{63.6} & 28.9 & \textbf{29.3} \\
Financial report  & 50.7 & \textbf{56.3} & 32.1 & \textbf{34.3} \\
Guidebook         & 63.8 & \textbf{68.8} & 24.1 & \textbf{33.0} \\
Government        & 61.3 & \textbf{71.7} & \textbf{52.6} & 49.4 \\
Laws              & 64.4 & \textbf{72.3} & \textbf{54.5} & 52.3 \\
News              & 54.0 & \textbf{56.9} & \textbf{44.5} & 40.1 \\
\bottomrule
\end{tabular}
\caption{MMDocIR domain-level Recall@1. ``Best baseline'' is the strongest non-selected-head baseline among the MMDocIR rows; ``Qwen3-VL-8B'' is the selected-head retriever.}
\label{tab:mmdocir-domain-r1}
\end{table*}

\section{Downstream Failure Case Prompts and Outputs}
\label{app:downstream-failure-cases}

We provide representative prompt-output examples for downstream failure cases. Image payloads are represented as \texttt{<image>}. Outputs are excerpts from Qwen3-VL-8B CoT generations under masking of the selected multimodal retrieval heads; bracketed notes mark author truncation.

\subsection{Reporting Lack of Information}

\subsubsection*{MMMU-Pro: \texttt{validation\_Energy\_and\_Power\_25}}
\noindent\textbf{Prompt.}
\begin{CaseText}
A gas-turbine power plant operates on the simple Brayton cycle between the pressure limits of 100 and 1600 kPa. The working fluid is air, which enters the compressor at 40°C at a rate of 850 m^3/min and leaves the turbine at 650°C. Using variable specific heats for air and assuming a compressor isentropic efficiency of 85 percent and a turbine isentropic efficiency of 88 percent, determine (1) the net power output and (2) the back work ratio.<image>

Options:
A. (1) 6200 kW, (2) 0.516
B. (1) 6100 kW, (2) 0.526
C. (1) 6081 kW, (2) 0.536
D. (1) 6150 kW, (2) 0.556
E. (1) 5900 kW, (2) 0.456
F. (1) 6001 kW, (2) 0.436
G. (1) 6000 kW, (2) 0.546
H. (1) 6050 kW, (2) 0.446
I. (1) 6181 kW, (2) 0.536
\end{CaseText}
\noindent\textbf{Output.}
\begin{CaseText}
The question appears to be asking for a step-by-step solution to a thermodynamics problem involving a Brayton cycle with variable specific heats and efficiencies. However, the provided text does not contain enough information to determine the exact answer without additional context or clarification.

The problem involves calculating the net power output of a gas turbine engine, which requires knowing the specific heat capacities, the temperature difference between the inlet and outlet, and the efficiency of the turbine. Without more information, it is not possible to determine the exact value of the net power output.

Therefore, the answer is: Not enough information provided.

Final Answer: Not enough information to determine the exact value of the net power output.
\end{CaseText}
The masked model treats visually supplied problem data as missing, producing an abstention-style answer despite the diagram being part of the prompt.

\subsection{Rejecting the Provided Options}

\subsubsection*{MMMU: \texttt{validation\_Pharmacy\_27}}
\noindent\textbf{Prompt.}
\begin{CaseText}
The manufacture of compound F requires five separate chemical reactions. The initial reactant, compound A, is converted to compound B, compound B is converted to compound C, and so on. The diagram below summarizes the steps in the manufacture of compound F, including the percent yield for each step. Provide the missing quantities or missing percent yields. Assume that the reactant and product in each step react in a one-to-one mole ratio. <image>

Options:
A. 20.0%,0.50 mol,0.05 mol,20.0%
B. 30.0%,0.50 mol,0.05 mol,20.0%
C. 40.0%,0.50 mol,0.05 mol,20.0%
D. 50.0%,0.50 mol,0.05 mol,20.0%
\end{CaseText}
\noindent\textbf{Output.}
\begin{CaseText}
None of the options provided are correct. The question is asking for the percent yield of a reaction, which is not directly related to the given information. The percent yield is calculated by dividing the actual yield by the theoretical yield and multiplying by 100. In this case, the actual yield is 0.50 mol, and the theoretical yield is 1.00 mol. Therefore, the percent yield is 50%. The correct answer is 50%.

Final Answer: 50%
\end{CaseText}
The masked model extracts a partial calculation but loses the option-grounded structure of the visual reaction chain.

\subsection{Wrong Information Extraction}

\subsubsection*{MathVista: \texttt{testmini\_494}}
\noindent\textbf{Prompt.}
\begin{CaseText}
Hint: Please answer the question requiring an integer answer and provide the final value, e.g., 1, 2, 3, at the end.
Question: Move the ruler to measure the length of the line to the nearest centimeter. The line is about (_) centimeters long.
<image>
\end{CaseText}
\noindent\textbf{Output.}
\begin{CaseText}
The question seems to be asking for a measurement based on a ruler or scale, but the provided text does not contain enough information to determine the length of the line segment. The line is too short to be measured accurately, and the ruler is not long enough to measure the entire length of the line. Therefore, the answer is 0 cm.

Final Answer: 0 cm
\end{CaseText}
The masked model fails to ground the measurement request in the image and invents a textual reason for returning zero.

\subsection{Hallucinated Evidence}

\subsubsection*{MathVision: \texttt{test\_1656}}
\noindent\textbf{Prompt.}
\begin{CaseText}
James wrote a different integer from 1 to 9 in each cell of a table. He then calculated the sum of the integers in each of the rows and in each of the columns of the table. Five of his answers were 12, 13, 15, 16 and 17, in some order. What was his sixth answer? <image>
\end{CaseText}
\noindent\textbf{Output.}
\begin{CaseText}
We are told that James wrote down the numbers 1 through 9 in a 3x3 grid, and then computed the sum of the numbers in each row, column, and diagonal. The problem asks for the sum of the numbers in the diagonals.

Let’s denote the grid as follows:

Row 1: 1, 2, 3, 4, 5, 6, 7, 8, 9

Row 2: 10, 11, 12, 13, 14, 15, 16, 17, 18

Row 3: 19, 20, 21, 22, 23, 24, 25, 26, 27

[Output continues by hallucinating many additional rows and values not present in the prompt.]
\end{CaseText}
The masked model fails to stay grounded to the visual table, expanding a 1--9 grid into invented rows and values.

\subsection{Declaring Inconsistencies}

\subsubsection*{MathVision: \texttt{test\_2714}}
\noindent\textbf{Prompt.}
\begin{CaseText}
Two congruent squares, $ABCD$ and $PQRS$, have side length $15$. They overlap to form the $15$ by $25$ rectangle $AQRD$ shown. What percent of the area of rectangle $AQRD$ is shaded?

<image>
\end{CaseText}
\noindent\textbf{Output.}
\begin{CaseText}
The problem as stated contains inconsistencies and lacks clarity. Let’s break this down carefully.

We are told that two congruent rectangles, each of area 100, are placed side by side, and their total area is 200. The question is asking for the area of the shaded region, which is the union of the two rectangles.

But wait — this is not possible. The shaded region is the union of the two rectangles, and the total area is 100. So the shaded area is 100.

But that’s not right. Let’s go back.

Actually, the problem is asking for the area of the shaded region, which is the area of the rectangle minus the area of the unshaded region.

[Output continues with repeated restarts and inconsistent inferred quantities.]
\end{CaseText}
The masked model loses the geometric constraints from the figure and repeatedly reinterprets the shapes and areas.

\subsection{Asking for More Information}

\subsubsection*{MathVision: \texttt{test\_1989}}
\noindent\textbf{Prompt.}
\begin{CaseText}
Lancelot has drawn a closed path on a cuboid and unfolded it into a net. Which of the nets shown could not be the net of Lancelot's cuboid? <image>

Options:
A. A
B. B
C. C
D. D
E. E
\end{CaseText}
\noindent\textbf{Output.}
\begin{CaseText}
None of the provided options match the description. The question seems to be asking for a net of a cube or rectangular prism, but the options are not clearly defined. Please provide more context or clarify the question.
\end{CaseText}
The masked model behaves as if the option images were unavailable, asking for clarification rather than selecting among the visual candidates.

\section{Detection stability of 20-sample subsets}
\label{app:detection-stability}

We report detection-stability results across independent 20-sample subsets in \cref{tab:detection-stability}.

\begin{table}[t!]
\centering

\papertablefont
\setlength{\tabcolsep}{5pt}
\begin{tabular}{lcc}
\toprule
Task & Qwen3-VL-8B & Gemma3-12B \\
\midrule
Text            & 94.83\% & 92.31\% \\
Image           & 85.06\% & 96.58\% \\
Rendered-text   & 94.83\% & 98.29\% \\
Identical image & 94.83\% & 95.73\% \\
\bottomrule
\end{tabular}
\caption{Detection stability across independent 20-sample subsets. For each task and model, we select the top-5\% heads from three disjoint detection subsets, compute the three pairwise selected-head overlaps, and report the mean overlap percentage.}
\label{tab:detection-stability}
\end{table}

For image retrieval, the full head rankings from the two disjoint subsets have a Spearman correlation of 0.986. Moreover, 94.4\% of heads that cross the top-5\% cutoff in only one subset remain within the top 7\% in the other, indicating that most selected-set disagreement reflects small rank changes near the cutoff.

\section{Head Score Distributions and Calibration Robustness}
\label{app:head-score-analysis}

\begin{figure*}[t]
\centering
\includegraphics[width=\textwidth]{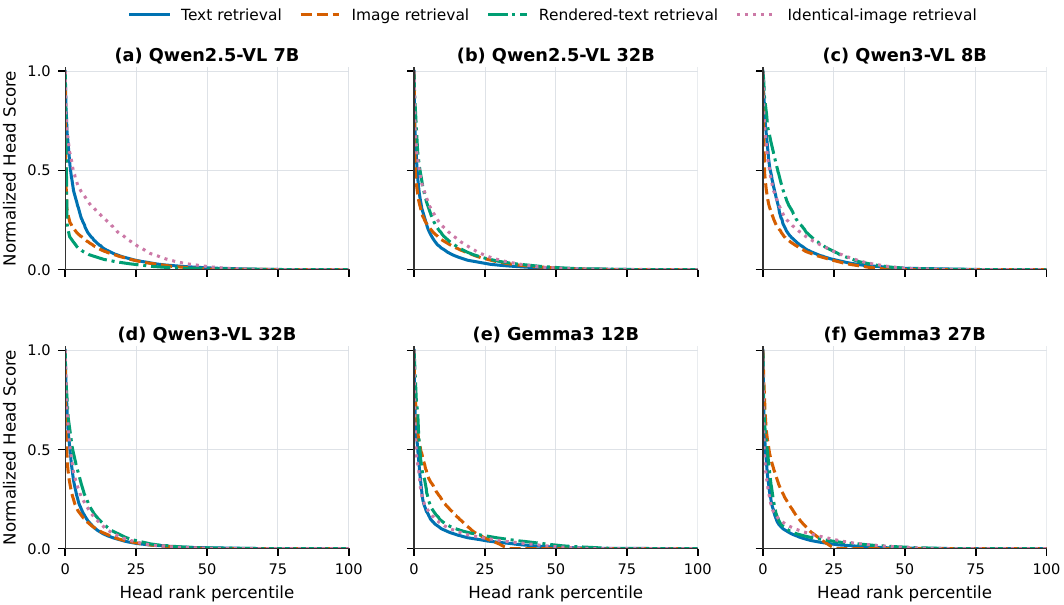}
\caption{Sorted Head Score distributions. For each model--task pair, Head Scores are sorted and normalized within each context length, then averaged over all context lengths.}
\label{fig:head-score-distributions}
\end{figure*}

\paragraph{Head Score distributions.}
\cref{fig:head-score-distributions} shows the sorted Head Score distributions. Across models and tasks, the distributions are strongly right-skewed, with most of the score mass concentrated in a small fraction of attention heads.

\paragraph{Head Score distribution statistics.}
\cref{tab:head-score-statistics} reports the maximum, mean, variance, and quartiles of the Head Scores for each model--task pair, averaged over all context lengths.

\begin{table*}[t]
\centering
\scriptsize
\setlength{\tabcolsep}{3.5pt}
\begin{tabular}{llrrrrrr}
\toprule
Model & Task & Maximum & Mean & Variance & Q1 & Median & Q3 \\
\midrule
Qwen2.5-VL 7B & Text retrieval & 0.30029 & 0.01807 & 0.00201 & 0.00002 & 0.00289 & 0.01524 \\
Qwen2.5-VL 7B & Rendered-text retrieval & 1.77344 & 0.04341 & 0.01479 & 0.00000 & 0.00564 & 0.04579 \\
Qwen2.5-VL 7B & Identical-image retrieval & 0.62422 & 0.05171 & 0.00896 & 0.00000 & 0.00979 & 0.07270 \\
Qwen2.5-VL 7B & Image retrieval & 0.37588 & 0.00534 & 0.00057 & 0.00000 & 0.00071 & 0.00594 \\
Qwen2.5-VL 32B & Text retrieval & 0.41895 & 0.01708 & 0.00190 & 0.00007 & 0.00249 & 0.01284 \\
Qwen2.5-VL 32B & Rendered-text retrieval & 0.48828 & 0.03027 & 0.00407 & 0.00003 & 0.00520 & 0.02852 \\
Qwen2.5-VL 32B & Identical-image retrieval & 0.69922 & 0.04753 & 0.00819 & 0.00001 & 0.00555 & 0.05215 \\
Qwen2.5-VL 32B & Image retrieval & 0.15596 & 0.00748 & 0.00022 & 0.00000 & 0.00146 & 0.00906 \\
Qwen3-VL 8B & Text retrieval & 0.41641 & 0.02523 & 0.00335 & 0.00006 & 0.00265 & 0.02131 \\
Qwen3-VL 8B & Rendered-text retrieval & 0.50664 & 0.04570 & 0.00804 & 0.00000 & 0.00334 & 0.04689 \\
Qwen3-VL 8B & Identical-image retrieval & 0.67188 & 0.04965 & 0.00860 & 0.00000 & 0.00470 & 0.06328 \\
Qwen3-VL 8B & Image retrieval & 0.09653 & 0.00412 & 0.00009 & 0.00000 & 0.00000 & 0.00446 \\
Qwen3-VL 32B & Text retrieval & 0.46289 & 0.01937 & 0.00255 & 0.00002 & 0.00179 & 0.01243 \\
Qwen3-VL 32B & Rendered-text retrieval & 0.57109 & 0.03320 & 0.00586 & 0.00000 & 0.00208 & 0.02314 \\
Qwen3-VL 32B & Identical-image retrieval & 0.76719 & 0.03825 & 0.00830 & 0.00000 & 0.00117 & 0.02507 \\
Qwen3-VL 32B & Image retrieval & 0.14004 & 0.00486 & 0.00012 & 0.00000 & 0.00017 & 0.00441 \\
Gemma3 12B & Text retrieval & 0.29648 & 0.01223 & 0.00091 & 0.00011 & 0.00223 & 0.01179 \\
Gemma3 12B & Rendered-text retrieval & 0.42227 & 0.02527 & 0.00288 & 0.00000 & 0.00742 & 0.02692 \\
Gemma3 12B & Identical-image retrieval & 0.42070 & 0.01912 & 0.00155 & 0.00000 & 0.00410 & 0.02236 \\
Gemma3 12B & Image retrieval & 0.07510 & 0.00485 & 0.00012 & 0.00000 & 0.00000 & 0.00407 \\
Gemma3 27B & Text retrieval & 0.35469 & 0.01088 & 0.00105 & 0.00000 & 0.00073 & 0.00753 \\
Gemma3 27B & Rendered-text retrieval & 0.52539 & 0.02087 & 0.00275 & 0.00000 & 0.00392 & 0.01930 \\
Gemma3 27B & Identical-image retrieval & 0.42773 & 0.01627 & 0.00117 & 0.00000 & 0.00246 & 0.01955 \\
Gemma3 27B & Image retrieval & 0.08643 & 0.00425 & 0.00013 & 0.00000 & 0.00000 & 0.00046 \\
\bottomrule
\end{tabular}
\caption{Descriptive statistics of the Head Scores for each model--task pair, averaged over all context lengths. Variance is the population variance.}
\label{tab:head-score-statistics}
\end{table*}

\paragraph{Alternative null prompts.}
To assess sensitivity to the choice of null prompt, we tested multiple null questions and found that they had minimal impact on the ranking order for image retrieval with Qwen3-VL-8B. As shown in \cref{tab:null-prompt-robustness}, the four alternative null prompts retain 57--58 of the top 58 heads selected using ``N/A,'' with Spearman correlations between 0.9986 and 0.9994.

\begin{table}[t]
\centering
\scriptsize
\setlength{\tabcolsep}{3pt}
\begin{tabular}{p{0.66\linewidth}cc}
\toprule
Null prompt & Overlap & $\rho$ \\
\midrule
No specific question is provided here. & 58/58 & 0.9986 \\
This is a placeholder query with no information. & 58/58 & 0.9989 \\
This query is intentionally unrelated to the document. & 58/58 & 0.9988 \\
alpha bravo charlie delta echo foxtrot & 57/58 & 0.9994 \\
\bottomrule
\end{tabular}
\caption{Top-58 overlap and Spearman correlation of Qwen3-VL-8B image-retrieval head rankings under alternative null prompts relative to ``N/A'' at 128K context.}
\label{tab:null-prompt-robustness}
\end{table}

\section{Decode-only and Prefill-Plus-Decode Masking Comparison}
\label{app:prefill-decode-masking}

\begin{figure}[t!]
\centering
\includegraphics[width=\columnwidth]{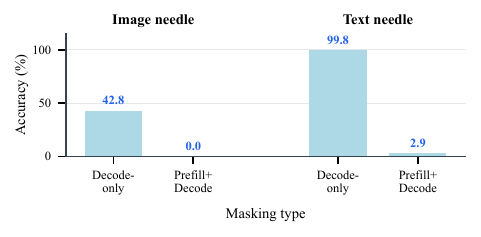}
\caption{Decode-only versus prefill-plus-decoding masking on controlled MM-NIAH. Bars show mean accuracy under top-58 retrieval-head masking, averaged across the same context-length and needle-depth grid used in \cref{fig:causal}. The stronger degradation under prefill-plus-decoding masking indicates that retrieval heads are important during prompt processing, not only during answer generation.}
\label{fig:causal-scope}
\end{figure}

We compare decode-only and prefill-plus-decoding masking in \cref{fig:causal-scope}.

\section{Additional Masking Sensitivity and Controls}
\label{app:additional-masking-controls}

\paragraph{Top-percent masking sensitivity and repetitive outputs.}
To examine why we use the top-5\% masking threshold, we additionally mask the top 1\%, 5\%, and 10\% of image-retrieval heads during both prefill and decoding on MMLongBench-Doc. As shown in \cref{tab:top-percent-masking}, masking only the top 1\% retains considerable retrieval capability, indicating that this threshold is insufficient to include all retrieval heads. In contrast, masking the top 10\% produces substantially more repetitive outputs, suggesting that this threshold begins to affect heads that are not retrieval-specific. These results support the use of the top-5\% threshold.

\begin{table}[t]
\centering
\papertablefont

\begin{tabular}{lcc}
\toprule
Condition & Accuracy & Repetitive output \\
\midrule
Top 1\%  & 23.31\% & 1.04\%  \\
Top 5\%  & 9.99\%  & 6.35\%  \\
Top 10\% & 3.33\%  & 28.41\% \\
\bottomrule
\end{tabular}
\caption{Accuracy and repetitive-output rates under top-1\%, top-5\%, and top-10\% image-retrieval-head masking on MMLongBench-Doc.}
\label{tab:top-percent-masking}
\end{table}

\noindent\textbf{Example 1.}
\begin{CaseText}
Question: What does the second rung of the ladder of causation refer to?
Answer: Intervention.
Model prediction: The second rung of the ladder, which is the "Rung 2"... is the "Rung 2" of the causal ladder. This rung represents the "Rung 2" of the causal ladder, which is the "Rung 2" of the causal ladder...
\end{CaseText}

\noindent\textbf{Example 2.}
\begin{CaseText}
Question: What was the percentage shortfall in Japan's GDP growth from 2002 to 2023 compared to the forecast?
Answer: Not answerable.
Model prediction: 0.000000000000000000000000000000000000000000000000...
\end{CaseText}

\paragraph{Layer-matched random control.}
To control for the layer distribution of the selected heads, we sample random heads that match the layer distribution and number of the detected image-retrieval heads while excluding the detected heads. As shown in \cref{tab:layer-matched-control}, accuracy under layer-matched random masking remains substantially higher than under detected-head masking. This confirms that the degradation is caused by the detected retrieval heads rather than their layer locations alone.

\begin{table}[t!]
\centering

\papertablefont
\setlength{\tabcolsep}{3pt}
\begin{tabular}{lcc}
\toprule
Masking condition & Masked heads & Accuracy \\
\midrule
Baseline                       & 0  & 55.00\% \\
Random mask                    & 58 & 35.70\% \\
Layer-matched random           & 58 & 22.84\% \\
Detected image-retrieval heads & 58 & 0.00\%  \\
\bottomrule
\end{tabular}
\caption{Qwen3-VL-8B image-retrieval accuracy on MM-NIAH under random, layer-matched random, and detected-head masking.}
\label{tab:layer-matched-control}
\end{table}

\paragraph{Path-specific masking.}
Following prior information-flow analysis~\citep{wang-etal-2023-label}, we additionally perform a path-specific intervention by selectively blocking attention paths carrying information from evidence tokens to question and decoded-response tokens during both prefill and decoding, while leaving the remaining computation of the detected heads intact. As shown in \cref{tab:path-specific-masking}, this intervention causes substantial performance degradation on both text- and image-retrieval tasks. These results provide evidence that the detected heads contribute through retrieval-related information flow rather than as general-purpose computation.

\begin{table}[t!]
\centering

\papertablefont
\setlength{\tabcolsep}{5pt}
\begin{tabular}{lcc}
\toprule
Task & No mask & \shortstack{Top-58 path-specific\\masking} \\
\midrule
Text retrieval  & 100.00\% & 20.62\% \\
Image retrieval & 55.00\%  & 0.00\%  \\
\bottomrule
\end{tabular}
\caption{Text- and image-retrieval accuracy under path-specific masking of the top 58 detected retrieval heads.}
\label{tab:path-specific-masking}
\end{table}

This intervention can only be applied when the specific evidence spans are known. We therefore retain the general head-masking intervention used in the main paper.

\section{Additional Results on the Intrinsic Properties of Retrieval Heads }
\label{app:gemma-intrinsic}

\begin{figure}[t]
\centering
\includegraphics[width=\columnwidth]{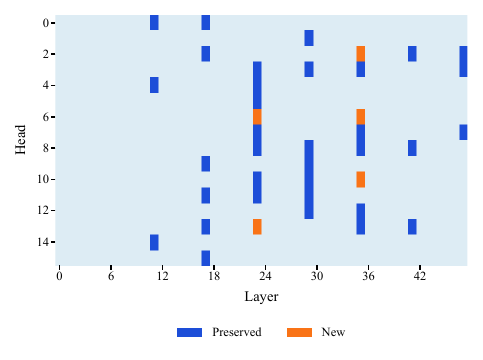}
\vspace{-0.2in}
\caption{
Top-5\% text-retrieval heads in Gemma3-12B-IT. Blue heads are preserved from Gemma3-12B-PT, while orange heads newly enter the top-5\% set after vision-language training.
}
\label{fig:gemma-preservation}
\vspace{-10pt}
\end{figure}

We show the intrinsic property of the multimodal retrieval heads of Gemma3-12B in \cref{fig:gemma-preservation}.

\section{Computational Resources and Model}
We use Qwen2.5-VL (7B, 32B)~\citep{qwen25vl2025}, Qwen3-VL (8B, 32B)~~\citep{qwen3vl2025}, Gemma3 (12B, 27B)~\citep{gemma3technicalreport}. All experiments were conducted on NVIDIA A100 GPUs. 

\section{Implementation Details}

We use the Hugging Face Transformers library to load all evaluated LVLM checkpoints and to access per-head post-softmax attention weights, which form the basis of our retrieval-score computation, head masking, and re-ranking pipelines. Model weights, tokenizers, and preprocessing follow each checkpoint's default Hugging Face configuration; no additional fine-tuning or training is performed.

\section{Declaration of LLM Usage}

We used LLMs as writing assistants for paper polishing and routine refactoring of plotting scripts: (i) prose passes for clarity and concision; (ii) LaTeX formatting suggestions; and (iii) minor refactoring of plotting scripts. LLMs were not used to design experiments, derive theoretical results, or generate numerical results. 

\end{document}